\documentclass{article} 
\usepackage{iclr2027_conference,times}

\usepackage{amsmath,amsfonts,bm}

\def\eqref#1{equation~\ref{#1}}

\def\1{\bm{1}}

\DeclareMathAlphabet{\mathsfit}{\encodingdefault}{\sfdefault}{m}{sl}
\SetMathAlphabet{\mathsfit}{bold}{\encodingdefault}{\sfdefault}{bx}{n}

\usepackage{hyperref}
\usepackage{url}
\usepackage{amsmath}
\usepackage{amssymb}
\usepackage{marvosym}
\usepackage{booktabs}
\usepackage{makecell}
\usepackage{multirow}
\usepackage{graphicx}
\usepackage{subcaption}
\usepackage{newfloat}
\usepackage{listings}
\usepackage{enumitem}

\definecolor{darkGreen}{RGB}{92, 148, 110}
\definecolor{asparagus}{RGB}{0.53, 0.66, 0.42}
\newcommand{\darkGreen}[1]{\textcolor{darkGreen}{#1}}

\usepackage[table]{xcolor} 
\newcommand{\method}{\textbf{RSI-Nav}}

\title{Data-Centric Recursive Self-Improvement\\
for Navigation via Policy-Synchronized\\
Language-Action Dual Bootstrapping}

\author{
\textbf{Songze Li}$^{*1,2}$ \qquad \textbf{Haoming Gu}$^{*3}$ \qquad \textbf{Ziyang Gong}$^{4}$ \qquad \textbf{Zhitong Wang}$^{5}$ \\
\textbf{Gengze Zhou}$^{6}$ \qquad \textbf{Zun Wang}$^{7}$ \qquad \textbf{Yuhan Zhu}$^{8}$ \qquad \textbf{Tianxiang Jiang}$^{2,9}$ \qquad \textbf{Juanzi Li}$^{5}$ \\
\textbf{Qi Wu}$^{6}$ \qquad \textbf{Weinan Zhang}$^{3}$ \qquad \textbf{Yu Qiao}$^{2}$ \qquad \textbf{Wenwei Zhang}\textsuperscript{\Letter\,2} \qquad \textbf{Yi Wang}\textsuperscript{\Letter\,2} \\
$^1$Fudan University \qquad $^2$Shanghai AI Laboratory \qquad $^3$Harbin Institute of Technology \\
$^4$Shanghai Jiao Tong University \qquad $^5$Tsinghua University \qquad $^6$University of Adelaide \\
$^7$UNC Chapel Hill \qquad $^8$Nanjing University \qquad $^9$University of Science and Technology of China \\
\vspace{0.1em}\\
{\tt \href{https://github.com/OpenGVLab/RSI-Nav}{https://github.com/OpenGVLab/RSI-Nav}}
}

\usepackage[most]{tcolorbox}

\newtcolorbox{senaprompt}[1]{
  enhanced,
  breakable,
  title={#1},
  fonttitle=\bfseries,
  fontupper=\normalsize,
  colback=white,
  colframe=gray!60,
  coltitle=black,
  boxrule=0.5pt,
  arc=3mm,
  width=0.98\linewidth,
  drop fuzzy shadow
}
\usepackage{float}
\usepackage{algorithm}
\usepackage{algorithmic}
\begin{document}

\maketitle

\begin{figure}[H]
  \centering
    \includegraphics[width=0.98\textwidth]{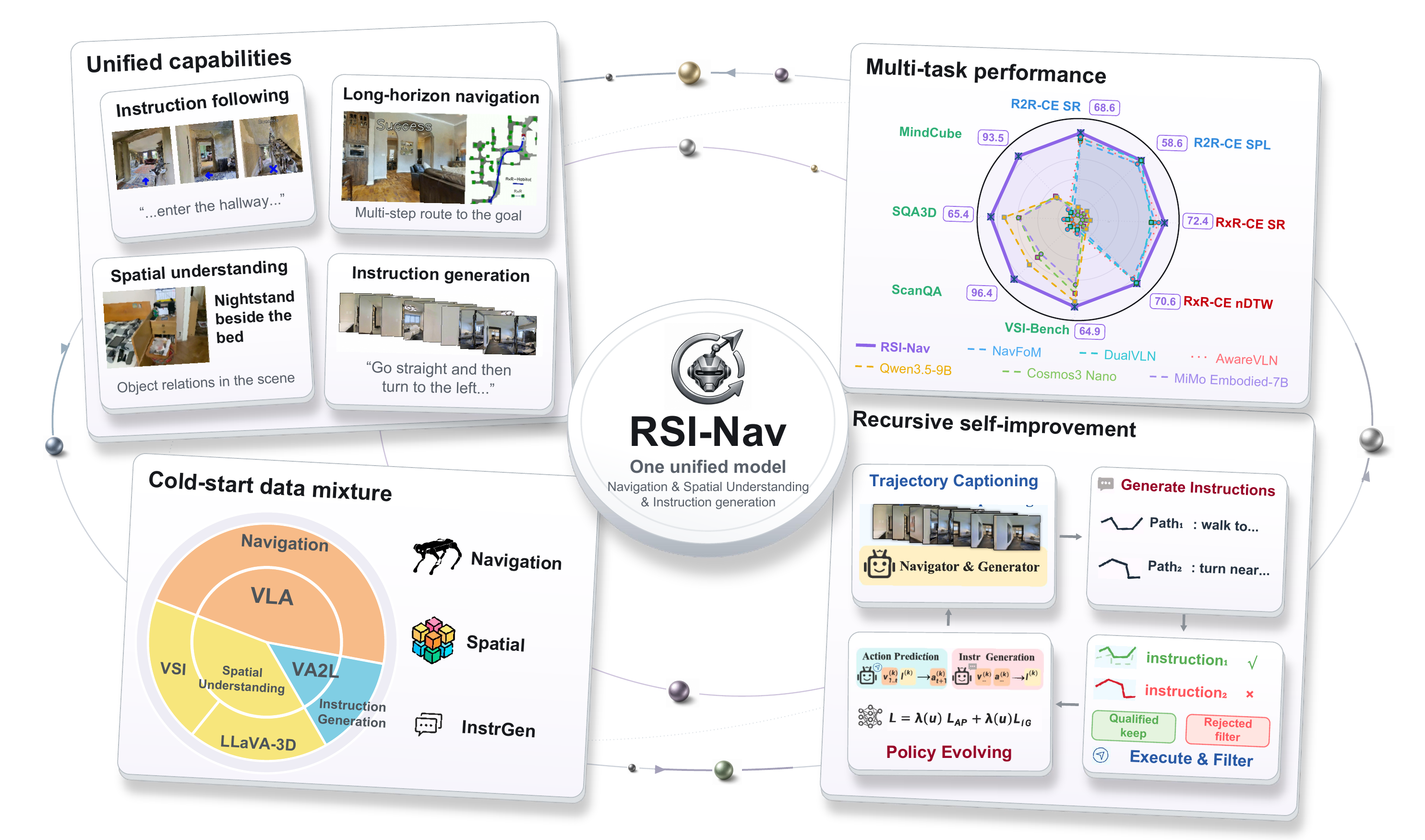}
    \caption{\method{} unifies navigation, instruction generation, and spatial understanding in one model, and recursively improves through policy-synchronized language-action dual bootstrapping.}
  \label{fig:teaser}
\end{figure}

\begin{abstract}
\begingroup
\renewcommand\thefootnote{}\footnotetext{%
*Equal contribution. \textsuperscript{\Letter} Corresponding authors.\\
\hspace*{1.8em}$\dagger$ This work was performed at Shanghai AI Laboratory.%
}%
\addtocounter{footnote}{-1}%
\endgroup

Vision-language navigation requires agents to follow natural-language instructions to reach target locations, making high-quality language supervision essential. Since human annotation is costly, prior work has explored scalable data synthesis methods. However, existing methods typically either use separate models for navigation and instruction generation or couple the two tasks through task-specific components; most also annotate the full trajectory corpus with a fixed model before updating, which can cause generated supervision to lag behind the evolving policy. Therefore, we introduce \textbf{RSI-Nav}, a data-centric recursive self-improvement framework for navigation via policy-synchronized language-action dual bootstrapping, which unifies vision-language-to-action instruction following and vision-action-to-language instruction generation within a shared model. Specifically, after supervised fine-tuning establishes fundamental capabilities, RSI-Nav progressively processes unlabeled trajectories in small shards. The current model generates instructions for these trajectories, executes them to evaluate goal completion and path efficiency, retains qualified samples, and immediately updates itself before processing the next shard. This shard-wise update keeps newly generated supervision aligned with the evolving policy. We also incorporate spatial understanding data, since stronger spatial reasoning benefits navigation. Extensive experiments show that RSI-Nav reaches 68.6\% SR on R2R-CE and 72.4\% SR on RxR-CE, while providing stronger instruction-generation utility and modestly improved spatial understanding after recursive evolution.
\end{abstract}

\section{Introduction}
\label{sec:intro}

Vision-language navigation (VLN) requires an embodied agent to ground natural-language instructions in visual observations and execute sequential actions to reach a target. This task begins with R2R~\citep{anderson2018r2r} and has expanded to longer-range navigation in RxR~\citep{anderson2020rxr}, goal-oriented settings such as REVERIE~\citep{qi2020reverie} and SOON~\citep{zhu2021soon}, and continuous navigation in VLN-CE~\citep{krantz2020beyond}. These increasingly challenging settings place greater demands on language grounding, spatial reasoning, and long-horizon decision making. Accordingly, research has progressed from improving navigation architectures and learning strategies~\citep{chen2021history,chen2022duet,wang2023gridmm,an2024etpnav} to scaling instruction-trajectory supervision through automatic data synthesis~\citep{chen2022hm3dlearning,wang2023scaling,wang2025bootstrapping}.

Scalable data synthesis allows navigation agents to learn from large amounts of unlabeled trajectories that would otherwise lack language supervision. Early methods use speaker models and back-translation to generate instructions for additional trajectories~\citep{fried2018speaker,tan2019learning}, while later work substantially scales this paradigm with larger environments and stronger instruction generators~\citep{chen2022hm3dlearning,wang2023scaling,wang2025bootstrapping,wang2025navrag}. Despite this progress, current pipelines still leave substantial room for improvement. Most methods learn instruction generation and navigation separately, while existing dual-learning approaches typically rely on task-specific architectures~\citep{wang2022counterfactual,wang2023lana}, limiting the benefit of bidirectional language-action learning. In addition, synthetic instructions are commonly generated by a fixed model, and the navigation policy is updated only after a large set of trajectories has been annotated. As the policy improves, the generated supervision can therefore become increasingly mismatched with the model being trained. These limitations call for a learning process that jointly exploits the two directions of language-action grounding while keeping newly generated supervision synchronized with the current policy.

This connection is particularly natural in navigation. More generally, complementary capabilities can benefit each other when one provides feedback for the other. An unlabeled trajectory can first be described as a route instruction and then executed again from its starting point, forming two complementary directions of language-action grounding. Instruction generation maps visual-action experience to language, while navigation maps language and visual observations back to actions. A faithful instruction should capture the navigation task well enough to guide the agent to the intended goal efficiently, making navigation execution a direct way to evaluate the generated supervision. When both capabilities are handled by the same evolving model, execution feedback can filter and improve language supervision, which in turn provides better data for subsequent navigation learning.

Based on this observation, we introduce \textbf{RSI-Nav}, a data-centric recursive self-improvement framework for navigation via policy-synchronized language-action dual bootstrapping. RSI-Nav uses a shared model to unify instruction following and generation for dual bootstrapping. We first apply supervised fine-tuning to establish basic capabilities for navigation and trajectory annotation. The model then progressively processes unlabeled trajectories in small shards. For each shard, the model generates instructions and executes them. The resulting rollouts assess whether generated instructions can guide the current model to the intended goal efficiently, and only qualified instruction-trajectory pairs are retained. The model is then immediately updated before the next shard. As a result, the updated model generates and evaluates the rollouts used in subsequent training, rather than relying on data produced by a fixed annotator at the beginning of the process. This shard-wise update keeps the generated supervision closely aligned with the evolving policy and forms a recursive process in which better models produce better training data for subsequent improvement.

Beyond language-action dual learning, we also incorporate spatial understanding into the same model, as spatial reasoning is crucial for interpreting routes and making navigation decisions~\citep{ma2022sqa3d,cheng2024spatialrgpt,yang2025thinking}. Joint training with spatial-understanding data strengthens this capability and in turn benefits navigation. We evaluate RSI-Nav on continuous vision-language navigation, trajectory-to-instruction generation, and spatial understanding. On R2R-CE~\citep{krantz2020beyond} and RxR-CE~\citep{anderson2020rxr}, RSI-Nav consistently improves navigation performance over its supervised initialization. 
The evolved model also maintains strong spatial understanding, with modest improvements on all four evaluated benchmarks over its supervised initialization, while its generated instructions provide stronger training utility for subsequent navigation. We further analyze the self-improvement process, examining policy synchronization, dual learning, action-conditioned instruction generation, and spatial supervision. These analyses explain why the proposed process improves both generated supervision and the navigation policy. Generally, our contributions can be summarized as follows:

\begin{itemize}[leftmargin=*]

\item We present \textbf{RSI-Nav}, a data-centric recursive self-improvement framework that unifies instruction following, instruction generation, and spatial understanding within a single model, where dual language-action learning provides complementary supervision to improve navigation.

\item We obtain large-scale execution-verified instructions paired with source trajectories as well as a strong unified model. To our knowledge, RSI-Nav is the first navigation framework to combine unified language-action dual learning with policy-synchronized recursive self-improvement.

\item We demonstrate strong navigation performance on VLN-CE benchmarks, including leading success rates among the compared methods, while maintaining strong spatial understanding. Analyses examine key components of the pipeline to explain how and why RSI-Nav improves.

\end{itemize}

\section{Related Work}
\label{sec:related_work}

\begin{figure*}
  \centering
    \includegraphics[width=0.98\textwidth]{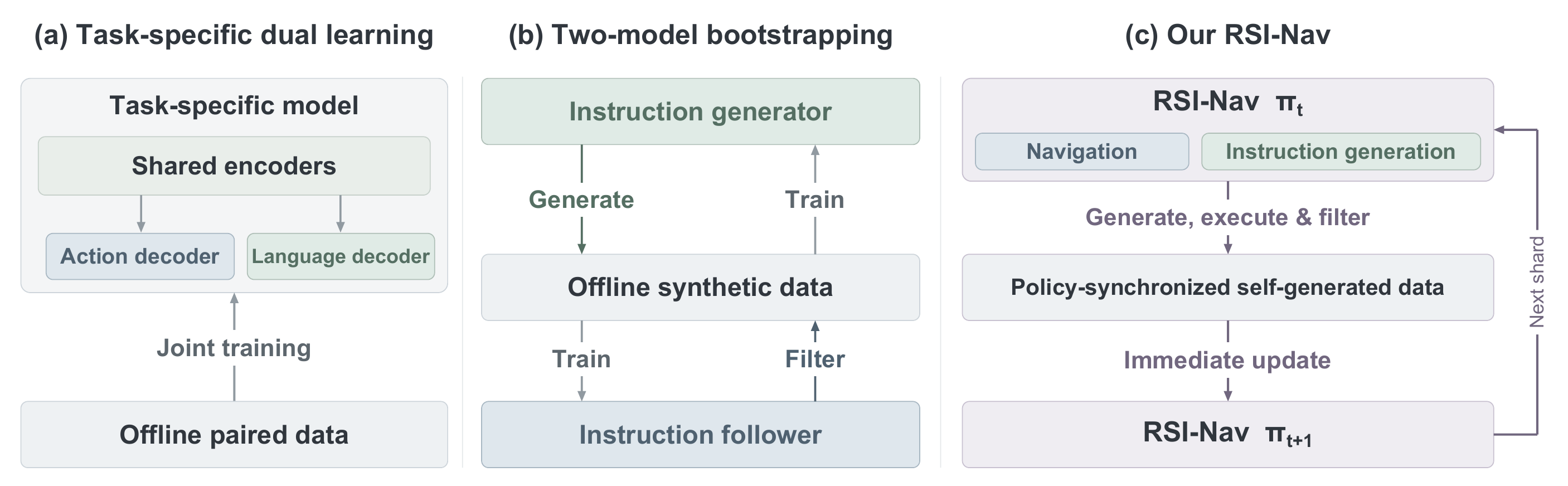}
    \caption{\textbf{Comparison of three VLN training paradigms.}
(a) Dual-task learning relies on task-specific architectures and costly human annotation.
(b) Data flywheel refines data at scale with separate navigation and instruction-generation models.
(c) RSI-Nav generates and filters data with the current policy and updates immediately after each shard, forming a policy-synchronized language-action dual bootstrapping loop for data-centric recursive self-improvement.}
  \label{fig:comp}
\end{figure*}

\paragraph{Vision-and-Language Navigation.}
Vision-and-Language Navigation (VLN) requires agents to follow natural-language instructions and navigate to target locations in visual environments. Existing scenarios include navigation with detailed action instructions~\citep{anderson2018r2r, anderson2020rxr, thomason2020vision}, remote object navigation with coarse-grained instructions~\citep{zhu2021soon, qi2020reverie}, and navigation in continuous environments~\citep{kolve2017ai2, krantz2020beyond, hong2022bridging}. Substantial progress has been made through vision-language pretraining and history modeling~\citep{hao2020towards,chen2021history,guhur2021airbert}, explicit map construction and planning~\citep{chen2022duet,wang2023gridmm,an2024etpnav}, and modern vision-language models that directly predict navigation actions~\citep{zhou2024navgpt2,zheng2024navillm,zhang2024navid}. However, the high cost of human annotation remains a major bottleneck, motivating automatic data augmentation and synthesis. Early methods use speaker models and back-translation~\citep{fried2018speaker,tan2019learning}, while later work exploits large-scale environments and synthetic instructions to construct millions of instruction-trajectory pairs~\citep{chen2022hm3dlearning,wang2023scaling,wang2025bootstrapping,li2025learning}. Recent studies further highlight spatial understanding, including situated 3D reasoning and spatial memory~\citep{yang2025thinking,yin2025spatial}.

\paragraph{Complementary Learning and Mutual Bootstrapping.}
Learning complementary capabilities together provides supervision and supports joint improvement across multiple abilities. In machine translation, dual learning couples forward and backward translation to provide richer mutual feedback~\citep{he2016dual}. Similar ideas also appear in mathematical reasoning, where a language model acts as both solver and verifier~\citep{weng2023large,lightman2023lets}, with recent work jointly training answer generation and self-verification in a single model~\citep{zhang2025selfverify}. These methods broadly share the principle that complementary capabilities can provide supervision or feedback for one another. In VLN, instruction following and generation form a similar bidirectional language-action relationship. Prior work realizes this through separate speaker and follower models, where generated instructions mainly serve as augmentation or scoring signals, or through task-specific architectures and consistency objectives~\citep{fried2018speaker,wang2022counterfactual,wang2023lana}.

\paragraph{Recursive Self-Improvement.}
Recursive self-improvement (RSI) broadly studies how an AI system can contribute to improving its subsequent versions. In the data-centric setting, RSIBench-Data~\citep{meng2026rsibenchdata} frames this as turning model failures and checkpoint feedback into better training data and, in turn, better models. Related approaches generate, filter, or verify model outputs for subsequent training in language reasoning~\citep{zelikman2022star,dong2023raft,xiong2025minimalist,yuan2024self} and embodied agents~\citep{bousmalis2023robocat,lin2025evolvenav}. In VLN, SRDF~\citep{wang2025bootstrapping} iteratively refines synthetic instructions, self-correction methods~\citep{yu2026correctnav} mine failure trajectories, and SID-VLN~\citep{li2025learning} reuses successful exploration trajectories. However, these methods generally separate data generation and model updates into relatively coarse rounds, allowing supervision to lag behind the evolving policy.
\section{Methodology}

\begin{figure*}[t]
  \centering
  \includegraphics[width=1\textwidth]{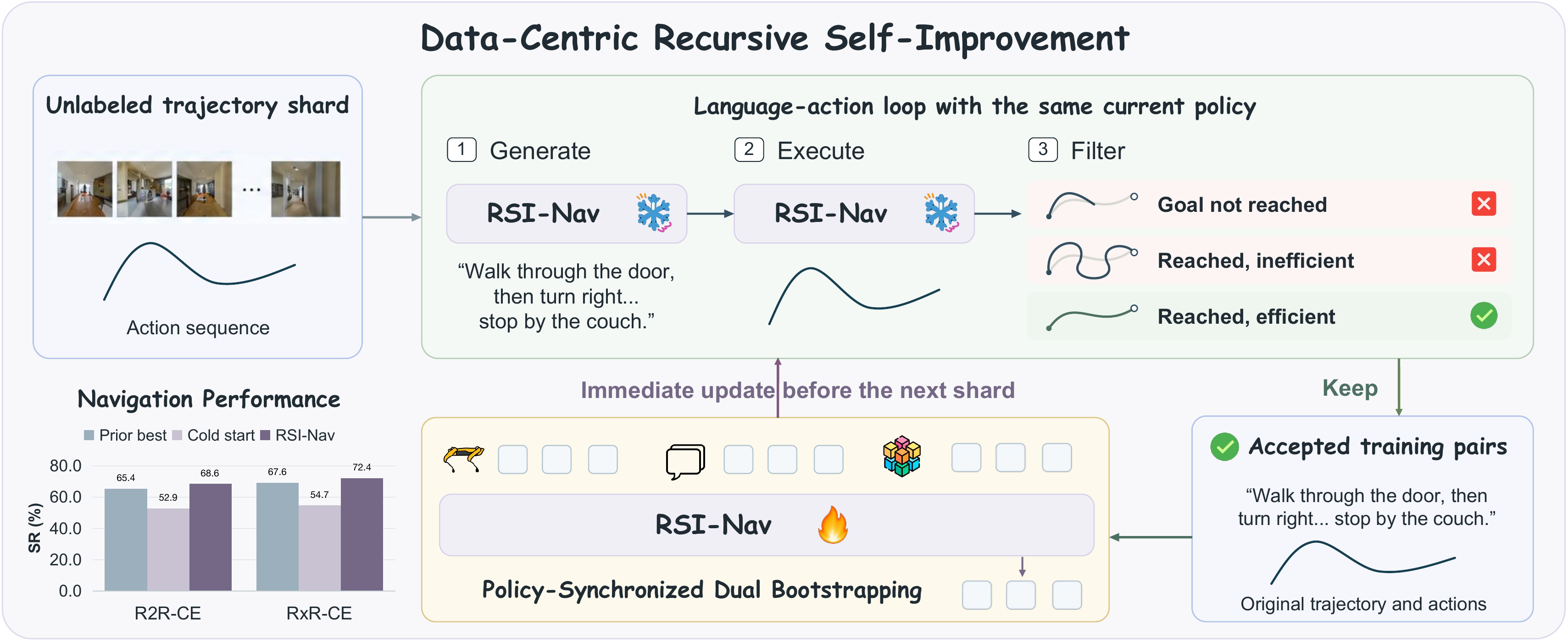}
  \caption{\textbf{Overview of RSI-Nav.}
  The current policy generates instructions for unlabeled trajectories and executes them to evaluate goal completion and efficiency. Qualified instructions are paired with the original trajectories to train instruction following and instruction generation jointly. The model is then updated with the accepted samples and spatial replay data before processing the next shard.}
  \label{fig:overall_pipeline}
\end{figure*}

RSI-Nav learns from unlabeled navigation trajectories through policy-synchronized language-action dual bootstrapping. As illustrated in Fig.~\ref{fig:overall_pipeline}, the framework begins with supervised initialization and recursively improves both the model and its generated supervision. We first introduce the continuous VLN setting, followed by a shared model that unifies instruction following, instruction generation, and spatial understanding in Sec.~\ref{sec:unified_model}. Sec.~\ref{sec:rsi} presents the core data-centric recursive self-improvement process, where the current policy generates instructions, executes them to assess goal completion and path efficiency, retains qualified samples, and updates itself before processing the next trajectory shard. Finally, Sec.~\ref{sec:training_data} describes how we curate supervision for initialization and recursive evolution, including spatial-data replay and newly generated instruction-trajectory pairs.

\paragraph{Preliminary.}
\label{sec:preliminary}

We consider vision-language navigation in continuous environments, where an agent follows a natural-language instruction through low-level interaction with the environment. At step $t$, the agent receives an egocentric RGB observation $o_t$ and selects an action
$a_t \in \mathcal{A}=\{\texttt{FORWARD},\texttt{LEFT},\texttt{RIGHT},\texttt{STOP}\}$.
A navigation trajectory is represented as
\begin{equation}
\tau_i =
\left(
o_1,a_1,o_2,a_2,\ldots,o_T,a_T
\right),
\end{equation}
where $T$ denotes the trajectory length. Each trajectory starts from an initial state $s_i^{\mathrm{start}}$ and terminates at a target location defined by its final state. Given a natural-language instruction $x_i$, the agent is required to reach the corresponding target efficiently. We use the term \emph{policy-synchronized} rather than strict on-policy learning. The source trajectories remain offline, while instruction generation and execution-based evaluation are repeatedly refreshed by the current model, reducing the gap between generated supervision and the evolving policy, as detailed in Sec.~\ref{sec:rsi}.

\subsection{A Unified Vision-Language-Action Model}
\label{sec:unified_model}

RSI-Nav is built on Qwen3.5-9B and uses a single model for instruction following, instruction generation, and spatial understanding. All capabilities share the same model parameters and are distinguished by task-specific prompts, without separate generators, verifiers, or output heads. We formulate the two complementary language-action tasks and present spatial understanding below.

\paragraph{Instruction Following.}
At navigation step $t$, the agent maintains an observation history
$H_t=\{o_1,\ldots,o_t\}$ and uniformly samples eight frames as visual context
$\widetilde{H}_t$. Conditioned on instruction $x$, $\widetilde{H}_t$, and the task prompt,
the model predicts a four-action chunk $c_t=(a_t,a_{t+1},a_{t+2},a_{t+3})$.
We optimize the standard cross-entropy loss over the predicted action chunk:
\begin{equation}
\mathcal{L}_{\mathrm{IF}}
=
-\log p_\theta(c_t\mid x,\widetilde{H}_t)
=
-\sum_{j=0}^{3}
\log p_\theta(a_{t+j}\mid x,\widetilde{H}_t,a_{t:t+j-1}).
\end{equation}
\paragraph{Instruction Generation.}
For the reverse direction, the model uniformly samples 32 frames
$\widetilde{V}_{\tau_i}$ from the trajectory and uses the full action sequence
$A_{\tau_i}=(a_1,\ldots,a_T)$ as input. Given these inputs and the task prompt,
the model generates $x=(w_1,\ldots,w_M)$ with the objective:
\begin{equation}
\mathcal{L}_{\mathrm{IG}}
=
-\log p_\theta(x\mid \widetilde{V}_{\tau_i},A_{\tau_i})
=
-\sum_{m=1}^{M}
\log p_\theta(w_m\mid \widetilde{V}_{\tau_i},A_{\tau_i},w_{<m}).
\end{equation}
\paragraph{Spatial Understanding.}
We further incorporate spatial-understanding training data, which provide an important scaffold for embodied decision making and navigation~\citep{gong2026acebrain0spatialintelligenceshared,brainteam2026acebrain05unifiedembodiedfoundational}. These samples are trained under the same token-level objective, denoted as $\mathcal{L}_{\mathrm{spa}}$.

\subsection{Policy-Synchronized Recursive Self-Improvement}
\label{sec:rsi}

With the shared instruction-following and instruction-generation model defined above, RSI-Nav recursively improves on a large corpus of unlabeled continuous trajectories. We partition the corpus into small, non-overlapping trajectory shards $\mathcal{U}=\{\mathcal{B}_1,\mathcal{B}_2,\ldots,\mathcal{B}_K\}$ and process each trajectory once. After each shard, the model is immediately updated before generating new supervision for the next one. This fine-grained update keeps data generation and execution-based evaluation close to the current policy, rather than relying on a fixed model over the entire corpus.

\paragraph{Generate, Execute, and Filter.}
For each trajectory $\tau_i\in\mathcal{B}_k$, the model generates an instruction $\hat{x}_i$ from the visual-action trajectory and then follows $\hat{x}_i$ from the original start state. The endpoint of $\tau_i$ serves as the navigation target. This execution tests whether the generated instruction can guide the current model to the intended destination efficiently. We retain a sample based on goal completion and path efficiency. An execution is considered successful if its final position is within 3\,m of the target. We further use SPL to measure navigation efficiency and define the acceptance indicator as
\begin{equation}
q_i =
\mathbb{I}
\left[
d_i < 3\text{ m}
\;\land\;
\operatorname{SPL}_i \geq \tau_{\mathrm{SPL}}
\right],
\end{equation}
where $d_i$ is the geodesic distance from the final position to the target and $\tau_{\mathrm{SPL}}$ is a fixed efficiency threshold. Only samples satisfying both criteria are retained for subsequent training.

\paragraph{Dual Bootstrapping and Shard-Wise Update.}
The filtering stage determines which generated instructions are reliable enough for training. For each accepted pair $(\tau_i,\hat{x}_i)$, we retain the original trajectory $\tau_i$. Since $\hat{x}_i$ is generated from $\tau_i$, successful and efficient execution toward the same goal provides an execution-grounded check that the instruction is executable and consistent with the navigation task represented by the source trajectory. We therefore retain the original trajectory as the supervision target rather than the executed trajectory, avoiding the accumulation of execution errors and the cost of repeatedly re-rendering continuous trajectories in Habitat. Keeping the trajectory fixed also focuses recursive improvement on the generated language supervision.

Each accepted pair provides supervision for both complementary directions. For instruction following, $\hat{x}_i$ is paired with the observation histories and original actions along $\tau_i$; for instruction generation, $(\widetilde{V}_{\tau_i},A_{\tau_i})$ is paired with $\hat{x}_i$. For shard $\mathcal{B}_k$, these accepted samples form $\mathcal{D}_k^{\mathrm{IF}}$ and $\mathcal{D}_k^{\mathrm{IG}}$, respectively. During the shard-wise update, we mix them with spatial-understanding data replayed from supervised initialization. The overall training objective is
\begin{equation}
\mathcal{L}_k
=
\mathcal{L}_{\mathrm{IF}}
\left(
\mathcal{D}_k^{\mathrm{IF}}
\right)
+
\mathcal{L}_{\mathrm{IG}}
\left(
\mathcal{D}_k^{\mathrm{IG}}
\right)
+
\mathcal{L}_{\mathrm{spa}}
\left(
\mathcal{R}_k^{\mathrm{spa}}
\right).
\end{equation}
Here, $\mathcal{L}_{\mathrm{IF}}$ and $\mathcal{L}_{\mathrm{IG}}$ follow the objectives defined in Sec.~\ref{sec:unified_model}, while $\mathcal{L}_{\mathrm{spa}}$ denotes the corresponding loss on spatial-understanding samples. This decomposition only distinguishes the three supervision sources; in practice, all samples are mixed and jointly optimized with a single SFT objective. The resulting $\theta_k$ is immediately used to process the next trajectory shard.

\paragraph{Recursive Self-Improvement.}
The updated model $\theta_k$ immediately processes the next shard $\mathcal{B}_{k+1}$. Refreshing instruction generation and execution-based filtering ensures that new supervision is produced by the latest model rather than a fixed policy. Repeating this cycle allows the model and its generated supervision to co-evolve across shards, while replayed spatial-understanding data help maintain the capability acquired during SFT initialization. The full evolution yields a stronger unified model and a large collection of execution-verified instruction-trajectory pairs, detailed in Sec.~\ref{sec:exp}.

\subsection{Training Data and Supervision}
\label{sec:training_data}

RSI-Nav is initialized with multi-source supervision for instruction following, instruction generation, and spatial understanding. We use continuous navigation data from SRDF~\citep{wang2025bootstrapping}, EnvDrop~\citep{tan2019learning}, ScaleVLN~\citep{wang2023scaling}, R2R~\citep{anderson2018r2r}, and RxR~\citep{anderson2020rxr}, and construct corresponding instruction-generation data by reversing the trajectory-instruction pairs. Since SRDF trajectories are originally discrete, we convert them into continuous trajectories for training, with details in Appendix~\ref{app:discrete-to-continuous-trajectory-conversion}. We also incorporate spatial-understanding data from VSI-590K~\citep{yang2026cambrian}, 546K samples from LLaVA-3D~\citep{zhu2024llava}, and MindCube~\citep{yin2025spatial}. All data are mixed for standard supervised fine-tuning, with the same spatial data replayed during recursive evolution. Starting from unlabeled continuous trajectories, RSI-Nav generates about 2.5M instruction-following samples and 1.1M instruction-generation samples through policy-synchronized recursive improvement.
\section{Experiments}
\label{sec:exp}

\begin{table*}[t]
    \centering
    {\fontsize{8pt}{9pt}\selectfont
    \setlength{\tabcolsep}{2.2pt}
    \caption{
    \textbf{Comparison on R2R-CE and RxR-CE Val-Unseen and four spatial-understanding benchmarks.}
    $^{*}$ denotes methods using the waypoint predictor from~\citet{hong2022bridging}.
    }
    \label{tab:r2r_rxr}
    \resizebox{\textwidth}{!}{
    \begin{tabular}{
        @{}l
        cccc
        @{\hspace{3pt}}cccc
        @{\hspace{3pt}}cccc
        @{\hspace{3pt}}cccc
        @{}
    }
    \toprule
    \multirow{2}{*}{Method}
    & \multicolumn{4}{c}{Observation}
    & \multicolumn{4}{c}{Spatial Understanding}
    & \multicolumn{4}{c}{R2R-CE Val-Unseen}
    & \multicolumn{4}{c}{RxR-CE Val-Unseen} \\
    \cmidrule(lr){2-5}
    \cmidrule(lr){6-9}
    \cmidrule(lr){10-13}
    \cmidrule(lr){14-17}

    & S.RGB & Pano. & Depth & Odo.
    & VSI-Bench$\uparrow$ & ScanQA$\uparrow$
    & SQA3D$\uparrow$ & MindCube$\uparrow$
    & NE$\downarrow$ & OSR$\uparrow$
    & SR$\uparrow$ & SPL$\uparrow$
    & NE$\downarrow$ & SR$\uparrow$
    & SPL$\uparrow$ & nDTW$\uparrow$ \\
    \midrule

    \multicolumn{17}{@{}l}{\darkGreen{\textit{Waypoint-Predictor-Based VLN Models}}} \\
    VLN$\circlearrowright$BERT$^{*}$~\citep{hong2022bridging}
    & & \checkmark & \checkmark & \checkmark
    & - & - & - & -
    & 5.74 & 53.0 & 44.0 & 39.0
    & 8.98 & 27.0 & 22.6 & 46.7 \\

    Sim2Sim$^{*}$~\citep{krantz2022sim}
    & & \checkmark & \checkmark & \checkmark
    & - & - & - & -
    & 6.07 & 52.0 & 43.0 & 36.0
    & - & - & - & - \\

    GridMM$^{*}$~\citep{wang2023gridmm}
    & & \checkmark & \checkmark & \checkmark
    & - & - & - & -
    & 5.11 & 61.0 & 49.0 & 41.0
    & - & - & - & - \\

    Reborn$^{*}$~\citep{an20221st}
    & & \checkmark & \checkmark & \checkmark
    & - & - & - & -
    & 5.40 & 57.0 & 50.0 & 46.0
    & 5.98 & 48.6 & 42.0 & 63.3 \\

    ETPNav$^{*}$~\citep{an2024etpnav}
    & & \checkmark & \checkmark & \checkmark
    & - & - & - & -
    & 4.71 & 65.0 & 57.0 & 49.0
    & 5.64 & 54.7 & 44.8 & 61.9 \\

    HNR$^{*}$~\citep{wang2024lookahead}
    & & \checkmark & \checkmark & \checkmark
    & - & - & - & -
    & 4.42 & 67.0 & 61.0 & 51.0
    & 5.50 & 56.3 & 46.7 & 63.5 \\

    BEVBert$^{*}$~\citep{an2023bevbert}
    & & \checkmark & \checkmark & \checkmark
    & - & - & - & -
    & 4.57 & 67.0 & 59.0 & 50.0
    & - & - & - & - \\

    SRDF$^{*}$~\citep{wang2025bootstrapping}
    & & \checkmark & \checkmark & \checkmark
    & - & - & - & -
    & 4.12 & - & 65.0 & 57.0
    & - & - & - & - \\

    \midrule

    \multicolumn{17}{@{}l}{\darkGreen{\textit{RGB-Based Continuous VLN Models}}} \\

    InstructNav~\citep{long2024instructnavzeroshotgenericinstruction}
    & & \checkmark & \checkmark & \checkmark
    & - & - & - & -
    & 6.89 & - & 31.0 & 24.0
    & - & - & - & - \\

    LAW~\citep{raychaudhuri2021language}
    & \checkmark & & \checkmark & \checkmark
    & - & - & - & -
    & 6.83 & 44.0 & 35.0 & 31.0
    & 10.90 & 8.0 & 8.0 & 38.0 \\

    Dynam3D~\citep{wang2026dynam3d}
    & \checkmark & & \checkmark & \checkmark
    & - & - & - & -
    & 5.34 & 62.1 & 52.9 & 45.7
    & - & - & - & - \\

    AO-Planner~\citep{chen2024affordances}
    & & \checkmark & \checkmark &
    & - & - & - & -
    & 5.55 & 59.0 & 47.0 & 33.0
    & 7.06 & 43.3 & 30.5 & 50.1 \\

    Seq2Seq~\citep{krantz2020beyond}
    & \checkmark & & \checkmark &
    & - & - & - & -
    & 7.77 & 37.0 & 25.0 & 22.0
    & 12.10 & 13.9 & 11.9 & 30.8 \\

    CMA~\citep{krantz2020beyond}
    & \checkmark & & \checkmark &
    & - & - & - & -
    & 7.37 & 40.0 & 32.0 & 30.0
    & - & - & - & - \\

    Uni-NaVid~\citep{uninavid}
    & \checkmark & & &
    & - & - & - & -
    & 5.58 & 53.3 & 47.0 & 42.7
    & 6.24 & 48.7 & 40.9 & - \\

    NaVILA~\citep{cheng2024navila}
    & \checkmark & & &
    & - & - & - & -
    & 5.22 & 62.5 & 54.0 & 49.0
    & 6.77 & 49.3 & 44.0 & 58.8 \\

    StreamVLN~\citep{streamvln}
    & \checkmark & & &
    & - & - & - & -
    & 4.98 & 64.2 & 56.9 & 51.9
    & 6.22 & 52.9 & 46.0 & 61.9 \\

    NavFoM~\citep{zhang2026embodied}
    & \checkmark & & &
    & - & - & - & -
    & 4.61 & 72.1 & 61.7 & 55.3
    & 4.74 & 64.4 & 56.2 & 65.8 \\

    JanusVLN~\citep{zeng2026janusvln}
    & \checkmark & & &
    & - & - & - & -
    & 4.78 & 65.2 & 60.5 & 56.8
    & 6.06 & 56.2 & 47.5 & 62.1 \\

    DualVLN~\citep{wei2026ground}
    & \checkmark & & &
    & - & - & - & -
    & 4.05 & 70.7 & 64.3 & 58.5
    & 4.58 & 61.4 & 51.8 & 70.0 \\

    AwareVLN~\citep{guo2026awarevln}
    & \checkmark & & &
    & - & - & - & -
    & 4.02 & 73.5 & 65.4 & 55.1
    & 3.95 & 67.6 & 56.1 & 65.7 \\
    
    \midrule

    \multicolumn{17}{@{}l}{\darkGreen{\textit{Spatial Understanding Models}}} \\
    RoboBrain2.5-8B~\citep{tan2026robobrain}
    & \checkmark & & &
    & 41.0 & 63.1 & 28.1 & 28.1
    & - & - & - & -
    & - & - & - & - \\

    MiMo Embodied-7B~\citep{hao2025mimo}
    & \checkmark & & &
    & 48.5 & 55.5 & 44.6 & 32.3
    & - & - & - & -
    & - & - & - & - \\
    
    Qwen3.5-9B~\citep{qwen3.5}
    & \checkmark & & &
    & 61.0 & 73.1 & 55.3 & 33.1
    & - & - & - & -
    & - & - & - & - \\
    
    Embodied R1.5~\citep{yuan2026embodied}
    & \checkmark & & &
    & 56.1 & 67.1 & 48.2 & 34.8
    & - & - & - & -
    & - & - & - & - \\
    
    Cosmos3 Nano~\citep{agarwal2026cosmos}
    & \checkmark & & &
    & 54.9 & 60.9 & 44.2 & 34.8
    & - & - & - & - 
    & - & - & - & - \\
    
    \midrule

    \rowcolor{gray!15}
    \textbf{\method{} Cold Start}
    & \checkmark & & &
    & 64.6 & 93.2 & 65.1 & 91.8
    & 5.54 & 66.4
    & 52.9 & 45.5
    & 5.39 & 54.7
    & 48.8 & 62.4 \\

    \rowcolor{gray!15}
    \textbf{\method{} (Ours)}
    & \checkmark & & &
    & \textbf{64.9} & \textbf{96.4} & \textbf{65.4} & \textbf{93.5}
    & \textbf{3.70} & \textbf{78.0}
    & \textbf{68.6} & \textbf{58.6}
    & \textbf{3.53} & \textbf{72.4}
    & \textbf{56.5} & \textbf{70.6} \\

    \bottomrule
    \end{tabular}
    }}
\end{table*}

\paragraph{Implementation Details.}
RSI-Nav is built on Qwen3.5-9B and trained in two stages on 64 GPUs. First, we perform supervised fine-tuning for one epoch on the multi-source data described in Sec.~\ref{sec:training_data}, using a learning rate of $5\times10^{-6}$, a global batch size of 128, and AdamW optimization. We then conduct recursive self-improvement on the unlabeled trajectory corpus, partitioned into non-overlapping shards of 16,384 trajectories. After each shard, qualified instruction-trajectory pairs are immediately used for model updating together with spatial-understanding replay before processing the next shard.

\paragraph{Benchmarks and Evaluation Metrics.}
We evaluate navigation on R2R-CE~\citep{krantz2020beyond} and RxR-CE~\citep{anderson2020rxr} using their validation-unseen splits. Following prior work, we report Navigation Error (NE), Oracle Success Rate (OSR), Success Rate (SR), and Success weighted by Path Length (SPL) on R2R-CE, and additionally nDTW on RxR-CE for trajectory alignment. For trajectory-to-instruction generation, we report BLEU-1, BLEU-4, METEOR, ROUGE, CIDEr, and SPICE. Since prior methods are mainly evaluated in discrete Matterport3D environments, whereas RSI-Nav operates in continuous VLN-CE with Habitat observations, these NLP scores serve as reference rather than strictly controlled comparisons. We further evaluate spatial understanding on VSI-Bench~\citep{yang2025thinking}, ScanQA~\citep{azuma2022scanqa}, SQA3D~\citep{ma2022sqa3d}, and MindCube~\citep{yin2025spatial}, reporting VSI Overall, ScanQA CIDEr, SQA3D EM1-R, and MindCube Accuracy, respectively. In the main comparison, we distinguish single-view RGB, panoramic RGB, depth, and odometry inputs, and separately mark methods using waypoint predictors.

\paragraph{Real-World Evaluation Setup.}
For real-world evaluation, we deploy RSI-Nav on a DEEP Robotics M20 robot. RSI-Nav runs on a remote server equipped with an NVIDIA GeForce RTX 4090 GPU and continuously processes egocentric RGB observations and language instructions to generate navigation actions for closed-loop execution. We evaluate the agent in both indoor and outdoor environments with tasks covering obstacle avoidance, narrow-space navigation, instruction following, open-vocabulary landmark grounding, and long-horizon navigation.

\begin{table*}[t]
\centering

\begin{minipage}[t]{0.49\textwidth}
\centering
\captionof{table}{
\textbf{Ablation study of SFT data composition.}
We examine the effects of dual learning, action conditioning, and spatial supervision.
}
\label{tab:data_composition}

{\fontsize{8pt}{9pt}\selectfont
\setlength{\tabcolsep}{2.2pt}
\begin{tabular*}{\linewidth}{
    @{\extracolsep{\fill}}
    cccc
    @{\hspace{4pt}}
    cccc
    @{}
}
\toprule
Nav & V2L & VA2L & Spatial
& NE$\downarrow$ & OSR$\uparrow$ & SR$\uparrow$ & SPL$\uparrow$ \\
\midrule
\checkmark &   &   &   & 6.19 & 57.2 & 44.7 & 38.6 \\
\checkmark & \checkmark &   &   & 5.78 & 58.1 & 47.0 & 38.8 \\
\checkmark &   & \checkmark &   & 5.69 & 59.9 & 48.8 & 41.0 \\
\checkmark &   & \checkmark & \checkmark
& \textbf{5.54} & \textbf{66.4} & \textbf{52.9} & \textbf{45.5} \\
\bottomrule
\end{tabular*}
}
\end{minipage}
\hfill
\begin{minipage}[t]{0.49\textwidth}
\centering
\captionof{table}{
\textbf{Ablation on policy update granularity.}
\emph{Candidate} denotes the number of trajectories processed per GPU before one policy update.
}
\label{tab:onpolicy_ablation}

{\fontsize{8pt}{9pt}\selectfont
\setlength{\tabcolsep}{2.2pt}
\begin{tabular*}{\linewidth}{
    @{\extracolsep{\fill}}
    lcccc
    @{}
}
\toprule
Candidate & NE$\downarrow$ & OSR$\uparrow$ & SR$\uparrow$ & SPL$\uparrow$ \\
\midrule
Cold Start & 5.54 & 66.4 & 52.9 & 45.5 \\
64       & 5.57 & 62.9 & 51.9 & 45.9 \\
256      & \textbf{4.91} & \textbf{68.7} & \textbf{57.2} & \textbf{48.7} \\
1024     & 5.03 & 68.1 & 55.1 & 46.3 \\
\bottomrule
\end{tabular*}
}
\end{minipage}
\end{table*}

\subsection{Main Results}

\paragraph{Comparison with Prior Methods.}
Tab.~\ref{tab:r2r_rxr} compares RSI-Nav with previous methods on R2R-CE, RxR-CE, and four spatial-understanding benchmarks. On navigation, RSI-Nav achieves strong performance on R2R-CE and RxR-CE Val-Unseen, reaching 68.6\% SR and 58.6\% SPL on R2R-CE, and 72.4\% SR, 56.5\% SPL, and 70.6 nDTW on RxR-CE. Compared with the supervised initialization, recursive evolution consistently improves navigation success and trajectory alignment, showing that self-generated and execution-verified instructions provide supervision beyond the initial SFT stage. The gain in nDTW indicates that the model better follows the instructed route rather than merely reaching the goal through alternative paths, which is particularly important for long-horizon instruction following. For spatial understanding, RSI-Nav maintains strong performance after evolution, with modest improvements over its cold-start checkpoint on all four benchmarks. VSI-Bench increases from 64.6 to 64.9, ScanQA from 93.2 to 96.4, SQA3D from 65.1 to 65.4, and MindCube from 91.8 to 93.5, suggesting that navigation gains coexist with stable spatial understanding.

\begin{figure*}[t]
    \centering
    \includegraphics[width=\textwidth]{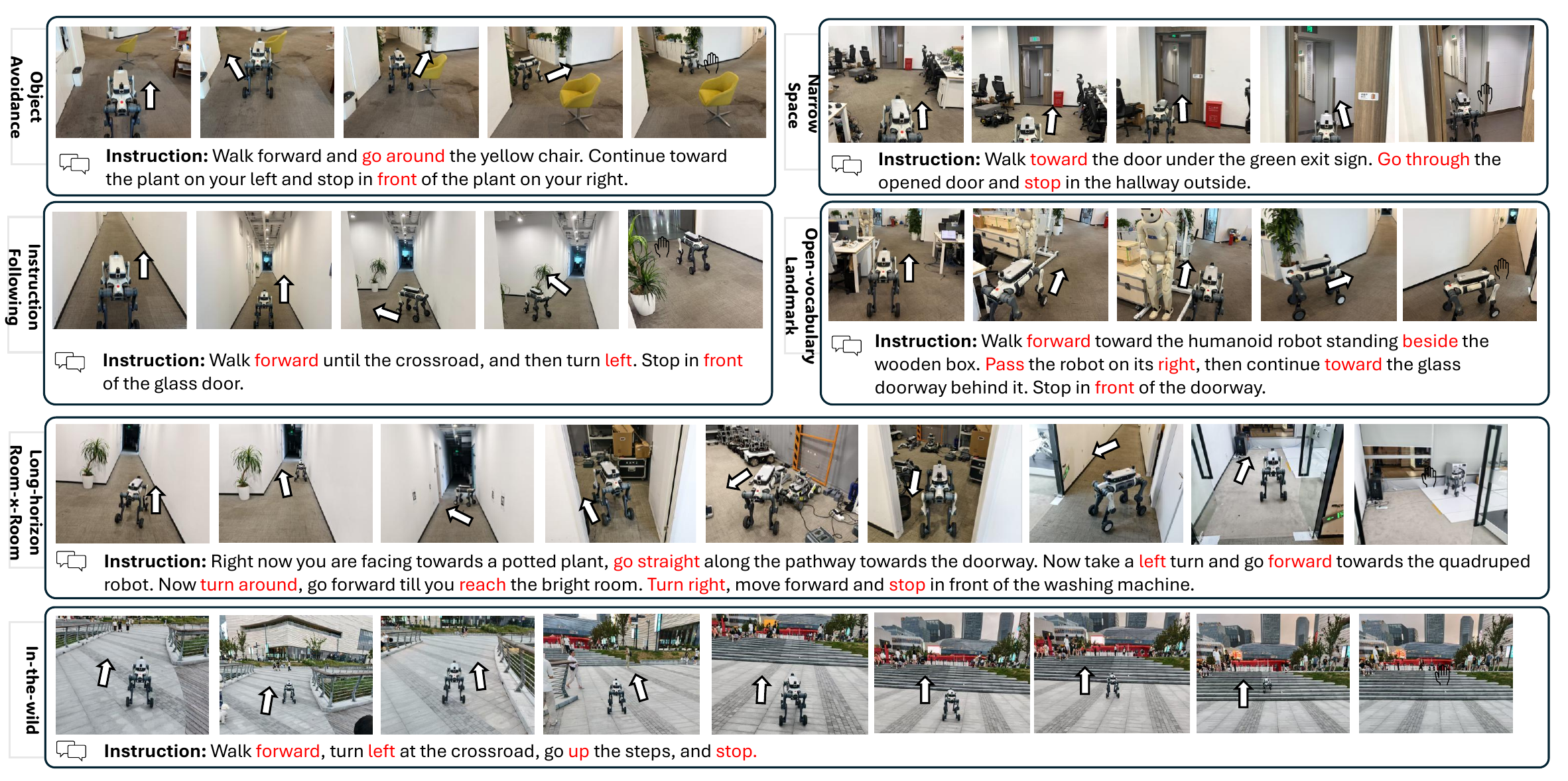}
    \caption{
    Qualitative results of RSI-Nav in real-world environments.
    }
    \label{fig:real_world}
\end{figure*}

\paragraph{Real-World Qualitative Results.}
Fig.~\ref{fig:real_world} presents qualitative results of RSI-Nav in real-world environments. RSI-Nav successfully handles diverse navigation scenarios, including obstacle avoidance, narrow-space traversal, multi-step instruction following, open-vocabulary landmark grounding, long-horizon room-x-room navigation, and in-the-wild outdoor navigation. These examples demonstrate successful closed-loop execution across several real-world scenarios.

\subsection{Ablation Studies}

\paragraph{How Does Policy Synchronization Affect Learning?}
We study the update granularity on a fixed subset of 64K unlabeled trajectories using 64 GPUs, where \emph{candidate} denotes the number of trajectories processed per GPU before each policy update. As shown in Tab.~\ref{tab:onpolicy_ablation}, 256 candidates perform best, improving SR and SPL most significantly. With only 64 candidates, performance also drops, suggesting that each update changes the policy too little for successive policies and their generated data to differ meaningfully. In contrast, 1,024 candidates process the entire subset before updating, approaching the conventional offline setting and enlarging the policy-data gap. The intermediate granularity therefore provides a better balance between meaningful policy evolution and timely supervision refresh. Based on this ablation, we use 256 candidates per GPU, corresponding to a global shard size of 16,384 trajectories on 64 GPUs, for full-scale recursive evolution.

\paragraph{Does Dual Learning Benefit Navigation?}
We examine whether instruction generation provides useful supervision for instruction following. As shown in Tab.~\ref{tab:data_composition}, adding VA2L data improves SR from 44.7\% to 48.8\% and SPL from 38.6\% to 41.0\% over navigation-only training, confirming that trajectory annotation benefits navigation. Tab.~\ref{tab:dual_learning} further shows that joint instruction following and generation substantially improves navigation while causing minor changes in instruction-generation metrics. This indicates that the two tasks can be jointly optimized with little loss in annotation quality, while the complementary trajectory-to-language supervision provides clear gains for navigation.

\paragraph{Are Actions Important for Dual Learning?}
Under the same visual input, VA2L consistently outperforms V2L, improving SR from 47.0\% to 48.8\% and SPL from 38.8\% to 41.0\%. Although instruction generation can rely on visual observations alone, incorporating actions establishes a stronger dual relation between action-to-language generation and language-to-action navigation. This indicates that action information helps more effectively transfer instruction-generation supervision back to navigation; we further examine this effect from the generation side next.

\begin{table*}[t]
\centering
{\fontsize{8pt}{9pt}\selectfont
\setlength{\tabcolsep}{2.2pt}

\caption{
\textbf{Trajectory-to-instruction generation and training utility.}
We report some NLP metrics for instruction generation and evaluate training utility by training navigation models on the generated supervision and measuring downstream R2R-CE navigation performance.
}
\label{tab:va2l_main}

\resizebox{\textwidth}{!}{
\begin{tabular}{
@{}l
cccccc
@{\hspace{5pt}}cccc
@{}
}
\toprule
\multirow{2}{*}{Method}
& \multicolumn{6}{c}{Instruction Generation}
& \multicolumn{4}{c}{Training Utility} \\
\cmidrule(lr){2-7}
\cmidrule(lr){8-11}

& BLEU-1$\uparrow$
& BLEU-4$\uparrow$
& METEOR$\uparrow$
& ROUGE$\uparrow$
& CIDEr$\uparrow$
& SPICE$\uparrow$
& NE$\downarrow$
& OSR$\uparrow$
& SR$\uparrow$
& SPL$\uparrow$ \\
\midrule

BT-speaker~\citep{fried2018speaker}
& 68.2 & 26.3 & 21.7 & 48.0 & 37.9 & 18.9
& - & - & - & - \\

EnvDrop~\citep{tan2019learning}
& 72.3 & 27.1 & 23.6 & 49.0 & 41.7 & 21.8
& - & - & - & - \\

CCC~\citep{wang2022counterfactual}
& 70.8 & 27.2 & 23.1 & 47.7 & 46.1 & 21.4
& - & - & - & - \\

KEFA~\citep{zeng2023kefa}
& 73.8 & 28.3 & 24.4 & 50.3 & 42.7 & 23.4
& - & - & - & - \\

LANA~\citep{wang2023lana}
& 73.6 & 28.9 & 23.7 & 49.8 & 45.7 & 22.6
& - & - & - & - \\

C-Instructor~\citep{kong2024controllable}
& 71.3 & 26.6 & 23.9 & 47.3 & 44.7 & 21.2
& - & - & - & - \\

BEVInstructor~\citep{fan2024navigation}
& 69.9 & 26.4 & 23.0 & 46.7 & 44.9 & 20.8
& - & - & - & - \\

SRDF~\citep{wang2025bootstrapping}
& 75.3 & 31.1 & 25.0 & 51.4 & 49.2 & 26.2
& 5.67 & 55.2 & 36.8 & 30.4 \\

\midrule
\rowcolor{gray!15}
\textbf{RSI-Nav}
& 71.0 & 28.2 & 23.2 & 48.5 & \textbf{50.6} & 20.1
& \textbf{5.58} & \textbf{58.7} & \textbf{41.3} & \textbf{35.1} \\

\bottomrule
\end{tabular}
}
}
\end{table*}

\begin{table*}[t]
\centering
{\fontsize{8pt}{9pt}\selectfont
\setlength{\tabcolsep}{2.2pt}

\caption{
\textbf{Ablation of bidirectional language--action learning.}
Joint VL2A--VA2L training substantially improves navigation while maintaining comparable instruction-generation quality.
}
\label{tab:dual_learning}

\resizebox{\textwidth}{!}{
\begin{tabular}{
@{}cc
@{\hspace{4pt}}cccc
@{\hspace{4pt}}cccccc
@{}
}
\toprule
\multicolumn{2}{c}{Training Objective}
& \multicolumn{4}{c}{Instruction Following (VL2A)}
& \multicolumn{6}{c}{Instruction Generation (VA2L)} \\
\cmidrule(lr){1-2}
\cmidrule(lr){3-6}
\cmidrule(lr){7-12}

VL2A & VA2L
& NE$\downarrow$
& OSR$\uparrow$
& SR$\uparrow$
& SPL$\uparrow$
& BLEU-1$\uparrow$
& BLEU-4$\uparrow$
& METEOR$\uparrow$
& ROUGE$\uparrow$
& CIDEr$\uparrow$
& SPICE$\uparrow$ \\
\midrule

\checkmark & 
& 6.19 & 57.2 & 44.7 & 38.6
& - & - & - & - & - & - \\

& \checkmark
& - & - & - & -
& 71.7 & 28.8 & 23.3 & 48.8 & 49.8 & 20.6 \\

\rowcolor{gray!15}
\checkmark & \checkmark
& \textbf{5.69}
& \textbf{59.9}
& \textbf{48.8}
& \textbf{41.0}
& 71.0
& 28.2
& 23.2
& 48.5
& \textbf{50.6}
& 20.1 \\

\bottomrule
\end{tabular}
}
}
\end{table*}

\paragraph{How Does Joint Training Affect Instruction Generation?} For RSI-Nav, we use the jointly trained VL2A-VA2L generator evaluated in Tab.~\ref{tab:dual_learning}. We examine the reverse benefit of dual learning from instruction following to instruction generation. Prior methods are often evaluated in discrete Matterport3D settings with panoramic observations, whereas RSI-Nav operates in continuous VLN-CE using low-resolution Habitat observations; their NLP scores thus serve only as reference rather than strictly comparable baselines. NLP metrics also mainly measure similarity to human-written instructions and may not reflect downstream training utility~\citep{gadre2024datacomp}. We therefore annotate the same trajectories with different generators and train identical navigation models on the resulting data. As shown in Tab.~\ref{tab:va2l_main}, although SRDF scores higher on most NLP metrics, RSI-Nav yields substantially better downstream navigation, improving SR from 36.8\% to 41.3\% and SPL from 30.4\% to 35.1\%. Together with the bidirectional-training results in Tab.~\ref{tab:dual_learning}, these results show that joint VL2A-VA2L training benefits navigation while preserving instruction-generation quality, and that the resulting RSI-Nav generator yields stronger downstream utility than prior work.

\paragraph{Does Spatial Understanding Benefit Navigation?}
Adding spatial-understanding supervision on top of VA2L further improves all navigation metrics, reducing navigation error while increasing success rate and efficiency. The gains support the role of spatial relations and viewpoint reasoning in navigation, highlighting spatial understanding as complementary supervision.

\paragraph{Limitations.} RSI-Nav still relies on simulated environments and an offline trajectory pool, while execution-based filtering incurs considerable computational cost. Although we provide qualitative real-world demonstrations, systematic robustness across diverse real-world environments, unseen domains, and substantially longer-horizon navigation remains to be established.
\section{Conclusion}
We introduce \textbf{RSI-Nav}, a data-centric recursive self-improvement framework for navigation via policy-synchronized language-action dual bootstrapping. RSI-Nav unifies instruction following, generation, and spatial understanding in one shared policy, turning unlabeled trajectories into supervision through instruction generation, execution-based filtering, and shard-wise updates. By refreshing supervision with the latest model, RSI-Nav maintains policy-synchronized learning while jointly training instruction following and generation. Experiments on R2R-CE and RxR-CE demonstrate strong navigation improvements, while spatial-understanding evaluations show modest gains over supervised initialization and generated instructions provide useful downstream navigation supervision. Ablations examine the effects of update granularity, dual-task training, action conditioning, and spatial supervision. Overall, RSI-Nav shows that large-scale unlabeled navigation trajectories support model improvement through generated and filtered supervision, providing a scalable direction for self-improving embodied navigation. We discuss limitations and future work in Appendix~\ref{app:limit}.

\section*{AI Use Disclosure}
We used generative AI tools only to assist with language polishing and editing of the manuscript, and with software programming. All AI-assisted text was reviewed and revised by the authors, and all AI-assisted code was inspected and tested for correctness. The authors take full responsibility for the final content of this work.

\bibliography{iclr2027_conference}
\bibliographystyle{iclr2027_conference}
\newpage
\section{Appendix}
\appendix
\setcounter{table}{0}
\renewcommand{\thetable}{A\arabic{table}}
\ifdefined\theHtable
  \renewcommand{\theHtable}{appendix.\arabic{table}}
\fi

\section{Evaluation Metrics}
\label{app:evaluation-metrics}

We evaluate navigation using NE, SR, OSR, SPL, and nDTW.
For episode $i$, let $P_i=(p_{i,0},\ldots,p_{i,T_i})$ denote
the executed trajectory, $g_i$ the goal, and
$R_i=(r_{i,1},\ldots,r_{i,M_i})$ the reference trajectory.
We use a success threshold of $\delta=3$ m.
Let $d_{\mathrm{geo}}$ denote geodesic distance on the
navigation mesh and $\mathbf{1}[\cdot]$ the indicator function.

\paragraph{Navigation Error (NE)} measures the geodesic distance between the final position and the goal:
\begin{equation}
\mathrm{NE}=\frac{1}{N}\sum_{i=1}^{N}
d_{\mathrm{geo}}(p_{i,T_i},g_i),
\end{equation}
where $N$ is the number of evaluation episodes.
Lower NE indicates that the agent finishes closer to the goal.

\paragraph{Success Rate (SR).}
An episode is successful if agent executes \texttt{STOP}
within the success threshold:
\begin{equation}
S_i=\mathbf{1}[\texttt{STOP}\text{ is executed}]
\mathbf{1}[d_{\mathrm{geo}}(p_{i,T_i},g_i)<\delta],
\qquad
\mathrm{SR}=\frac{1}{N}\sum_{i=1}^{N}S_i.
\end{equation}

\paragraph{Oracle Success Rate (OSR).}
OSR measures whether the agent enters the success region
at any point during navigation, regardless of where it stops:
\begin{equation}
\mathrm{OSR}=\frac{1}{N}\sum_{i=1}^{N}
\mathbf{1}\!\left[
\min_{0\leq t\leq T_i}
d_{\mathrm{geo}}(p_{i,t},g_i)<\delta
\right].
\end{equation}

\paragraph{Success weighted by Path Length (SPL).}
SPL considers both success and navigation efficiency:
\begin{equation}
\mathrm{SPL}=\frac{1}{N}\sum_{i=1}^{N}
S_i\frac{\ell_i}{\max(\ell_i,L_i)},
\end{equation}
where $\ell_i=d_{\mathrm{geo}}(p_{i,0},g_i)$ is the shortest
start-to-goal distance and
$L_i=\sum_{t=1}^{T_i}\|p_{i,t}-p_{i,t-1}\|_2$
is the traveled distance. Failed episodes receive zero,
while successful episodes are penalized for taking detours.

\paragraph{Normalized Dynamic Time Warping (nDTW).}
nDTW measures how closely the executed trajectory follows
the reference trajectory while preserving traversal order.
The alignment cost is
\begin{equation}
\mathrm{DTW}(P_i,R_i)
=
\min_{W\in\mathcal{W}_i}
\sum_{(t,j)\in W}\|p_{i,t}-r_{i,j}\|_2,
\end{equation}
where $\mathcal{W}_i$ contains monotonic alignments connecting
the starts and ends of both trajectories, with each step
advancing along either or both sequences.
The normalized score is
\begin{equation}
\mathrm{nDTW}
=
\frac{1}{N}\sum_{i=1}^{N}
\exp\left(
-\frac{\mathrm{DTW}(P_i,R_i)}{M_i\delta}
\right).
\end{equation}
Our implementation removes consecutive duplicate predicted
positions and uses FastDTW by default to approximate the
alignment cost. Higher nDTW indicates better agreement
with the reference route.

NE is reported in meters. SR, OSR, SPL, and nDTW are
reported on a percentage scale, with higher values
indicating better performance.

\section{Action Representation}
\label{app:action-representation}

We compare three action representations: \textbf{Arrow} uses $\uparrow/\leftarrow/\rightarrow$ and \texttt{stop}, \textbf{Single} uses \texttt{forward/left/right/stop}, and \textbf{Double} uses \texttt{move forward/turn left/turn right/stop}.

\begin{table}[!htbp]
\centering
\small
\setlength{\tabcolsep}{4pt}
\caption{Action representation ablation on R2R val-unseen.}
\label{tab:appendix-1}
\begin{tabular}{lrrrr}
\toprule
Representation & NE \(\downarrow\) & OSR \(\uparrow\) & SR \(\uparrow\) & SPL \(\uparrow\) \\
\midrule
Arrow & 6.32 & 51.22 & 43.28 & 37.80 \\
Single & \textbf{5.56} & \textbf{57.48} & \textbf{48.72} & \textbf{42.67} \\
Double & 5.92 & 53.78 & 43.83 & 39.22 \\

\bottomrule
\end{tabular}
\end{table}

Table~\ref{tab:appendix-1} shows that Single achieves the best
performance across all four metrics. Compared with Arrow, it
improves SR and SPL by 5.44 and 4.87 percentage points, respectively,
and also outperforms Double. These results support using
single-word action representations in our navigation model.

\section{Action Inputs and Trajectory Sampling}
\label{app:action-inputs-and-trajectory-sampling}

We compare \textbf{V2L}, which generates instructions from images alone, with \textbf{VA2L}, which also uses actions. We consider two frame sampling strategies: \textbf{Uniform-32} uniformly samples up to 32 frames across the trajectory, while \textbf{Interval-4} samples one frame every four action steps.

\begin{table}[!htbp]
\centering
\small
\setlength{\tabcolsep}{4pt}
\caption{Downstream navigation results with different action inputs and frame sampling strategies.}
\label{tab:appendix-2}
\begin{tabular}{lllrrrr}
\toprule
Representation & Supervision & Sampling & NE $\downarrow$ & OSR $\uparrow$ & SR $\uparrow$ & SPL $\uparrow$ \\
\midrule
Arrow & V2L & Interval-4 & 6.39 & 48.23 & 41.44 & 36.27 \\
Arrow & VA2L & Interval-4 & 6.20 & 47.14 & 41.49 & 37.62 \\
Arrow & V2L & Uniform-32 & 6.41 & 48.67 & 42.09 & 36.82 \\
Arrow & VA2L & Uniform-32 & 5.71 & 50.68 & 44.15 & 39.54 \\
\midrule
Single & V2L & Interval-4 & 5.40 & 54.98 & 47.25 & 40.59 \\
Single & VA2L & Interval-4 & 5.42 & 55.36 & 48.02 & 41.36 \\
Single & V2L & Uniform-32 & 5.15 & 58.34 & 51.22 & 42.88 \\
Single & VA2L & Uniform-32 & \textbf{5.00} & \textbf{60.14} & \textbf{53.13} & \textbf{45.38} \\
\bottomrule
\end{tabular}
\end{table}

Table~\ref{tab:appendix-2} shows that Uniform-32 consistently
improves SR and SPL over Interval-4 across both action
representations and supervision settings. Adding action inputs
through VA2L further improves these two metrics in all settings.
Combining Single, VA2L, and Uniform-32 achieves the best overall
performance, with an SR of 53.13 and an SPL of 45.38.

\section{Discrete-to-Continuous Trajectory Conversion}
\label{app:discrete-to-continuous-trajectory-conversion}

We convert discrete navigation paths into executable trajectories in Habitat. Given a path $\tau^G=(v_1,\ldots,v_N)$, we query MatterSim for each viewpoint's position $p_i^G=(x_i,y_i,z_i)$ and map it to Habitat coordinates:
\begin{equation}
\tilde{w}_i=(x_i,z_i-1.25,-y_i),
\qquad
w_i=(x_i,h_i,-y_i),
\label{eq:coordinate-conversion}
\end{equation}
where $1.25$ m is the camera-height offset and $h_i$ is the floor height obtained by snapping $\tilde{w}_i$ to the navigation mesh. If snapping fails, we use $h_i=z_i-1.25$.

Starting from the first waypoint with the converted initial orientation, a shortest-path follower follows the waypoints in order, producing
\begin{equation}
\tau^G
\xrightarrow{\text{waypoint following}}
\tau^C=(s_0,a_1,s_1,\ldots,a_L,s_L),
\label{eq:trajectory-conversion}
\end{equation}
where $s_l$ denotes the agent's position and orientation after action $a_l$, and $L$ is the number of executed actions. Each movement action advances the agent by $0.25$ m or turns it left or right by $15^\circ$. Trajectories that exhaust the action budget are discarded. The recorded actions and rendered observations are used to construct navigation and instruction-generation examples.

\begin{algorithm}[t]
\caption{Discrete-to-Continuous Trajectory Conversion}
\label{alg:de2ce}
\begin{algorithmic}[1]
\REQUIRE Discrete path $\tau^G=(v_1,\ldots,v_N)$, initial
orientation, navigation mesh $\mathcal{M}$, action budget $B=500$
\ENSURE Habitat waypoints $W$, initial state $s_0$, and action
sequence $A$, or rejection

\STATE $W \leftarrow [\,]$, $A \leftarrow [\,]$
\FOR{$i=1,\ldots,N$}
    \STATE Query MatterSim for $(x_i,y_i,z_i)$ at $v_i$
    \STATE $\tilde{w}_i \leftarrow (x_i,z_i-1.25,-y_i)$
    \STATE $q_i \leftarrow \operatorname{Snap}_{\mathcal{M}}
    (\tilde{w}_i)$
    \STATE $w_i \leftarrow \tilde{w}_i$
    \IF{$q_i$ is valid}
        \STATE $w_i \leftarrow (x_i,[q_i]_y,-y_i)$
        \COMMENT{Adjust height only}
    \ENDIF
    \STATE Append $w_i$ to $W$
\ENDFOR

\IF{$d_{\mathcal{M}}(w_1,w_N) \geq 100\,\mathrm{m}$}
    \RETURN rejection
\ENDIF

\STATE Initialize Habitat at $w_1$ with the converted initial
orientation; record $s_0$
\STATE Initialize shortest-path follower $F$ with goal radius
$0.25\,\mathrm{m}$
\FOR{$i=2,\ldots,N$}
    \LOOP
        \IF{$|A| \geq B$}
            \RETURN rejection
        \ENDIF
        \STATE $a \leftarrow F.\operatorname{NextAction}(w_i)$
        \IF{$a=\texttt{STOP}$ or $a=\texttt{None}$}
            \STATE \textbf{break}
        \ENDIF
        \STATE Execute $a$ in Habitat and append $a$ to $A$
    \ENDLOOP
\ENDFOR
\RETURN $(W,s_0,A)$
\end{algorithmic}
\end{algorithm}

\section{Prompt Templates}
\label{app:prompt-templates}

\subsection{Navigation}

The navigation prompt includes up to eight history frames and the current
view. The history clause is omitted when no history is available.
The model predicts up to four actions per turn; the responses below
illustrate the dialogue format.

\begin{senaprompt}{Navigation Prompt}
\textbf{User:} You are an autonomous navigation assistant. Your task is to
\{instruction\}. Devise an action sequence to follow the instruction using
the four actions: TURN LEFT (left) or TURN RIGHT (right) by 15 degrees,
MOVE FORWARD (forward) by 25 centimeters, or STOP (stop).

\textcolor{gray}{\textit{These are your historical observations:
\{history\_images\}}}

This is your current view: \textless image\textgreater

\textbf{Assistant:} forward,forward,left,forward

\textbf{User:} in front of you is \textless image\textgreater

\textbf{Assistant:} forward,stop
\end{senaprompt}

\subsection{Instruction Generation}

VA2L uses the following system and user messages:

\begin{senaprompt}{Instruction Generation Prompt}
\textbf{System:} You are a navigation-instruction generator.
Output only the requested instruction.

\textbf{User:} This is a navigation trajectory consisting of several image
sequences:

(Viewpoint 1: Image:\textless image\textgreater,
Action: \{action\_span\_1\}),

\ldots

(Viewpoint K: Image:\textless image\textgreater,
Action: \{action\_span\_K\}).

Could you give me its corresponding navigation instruction in details?

\textbf{Assistant:} \{navigation\_instruction\}
\end{senaprompt}

The viewpoint list is expanded for all sampled frames. Each action span
is described using phrases such as \texttt{forward 2 steps, then left}.
V2L removes the action fields.

\section{Visualization of Spatial Question Answering}
\label{app:spatial-qa-visualization}

\begin{figure}[!htbp]
    \centering
    \includegraphics[
        width=\linewidth,
        height=25\baselineskip,
        keepaspectratio
    ]{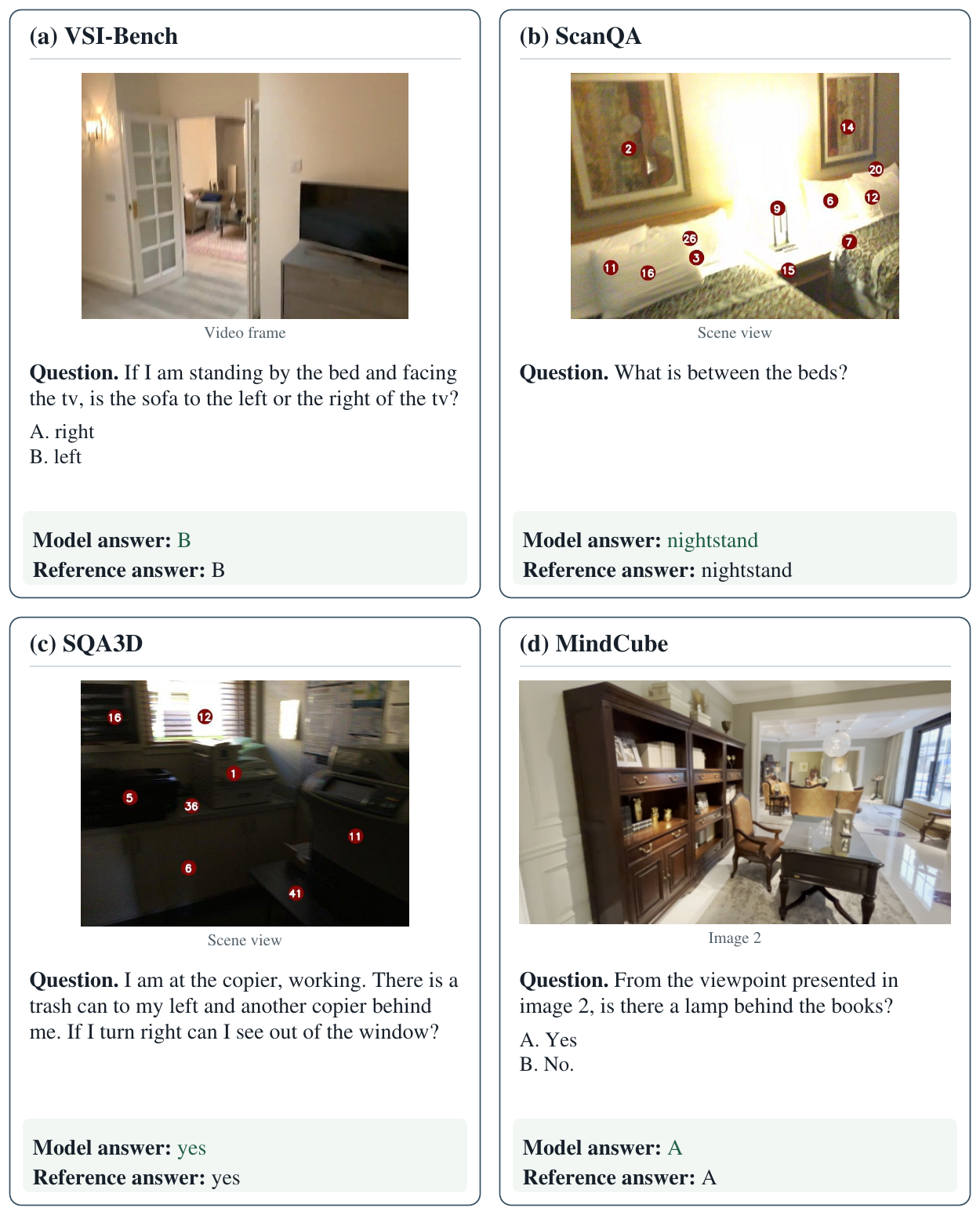}
    \caption{Selected correct QA examples on VSI-Bench,
    ScanQA, SQA3D, and MindCube.}
    \label{fig:spatial-qa}
\end{figure}

Figure~\ref{fig:spatial-qa} presents selected correct predictions
on VSI-Bench, ScanQA, SQA3D, and MindCube. These examples cover
relative direction reasoning, object identification, situated
question answering, and viewpoint-dependent spatial relations.
Each example includes one representative image, the question,
and the model answer alongside a matching reference answer.

\clearpage
\section{Visualization of R2R-CE and RxR-CE}
\label{app:navigation-visualization}

\begingroup
\setlength{\intextsep}{3pt}
\begin{figure}[H]
    \centering
    \includegraphics[
        width=0.8\linewidth,
        height=0.65\textheight,
        keepaspectratio
    ]{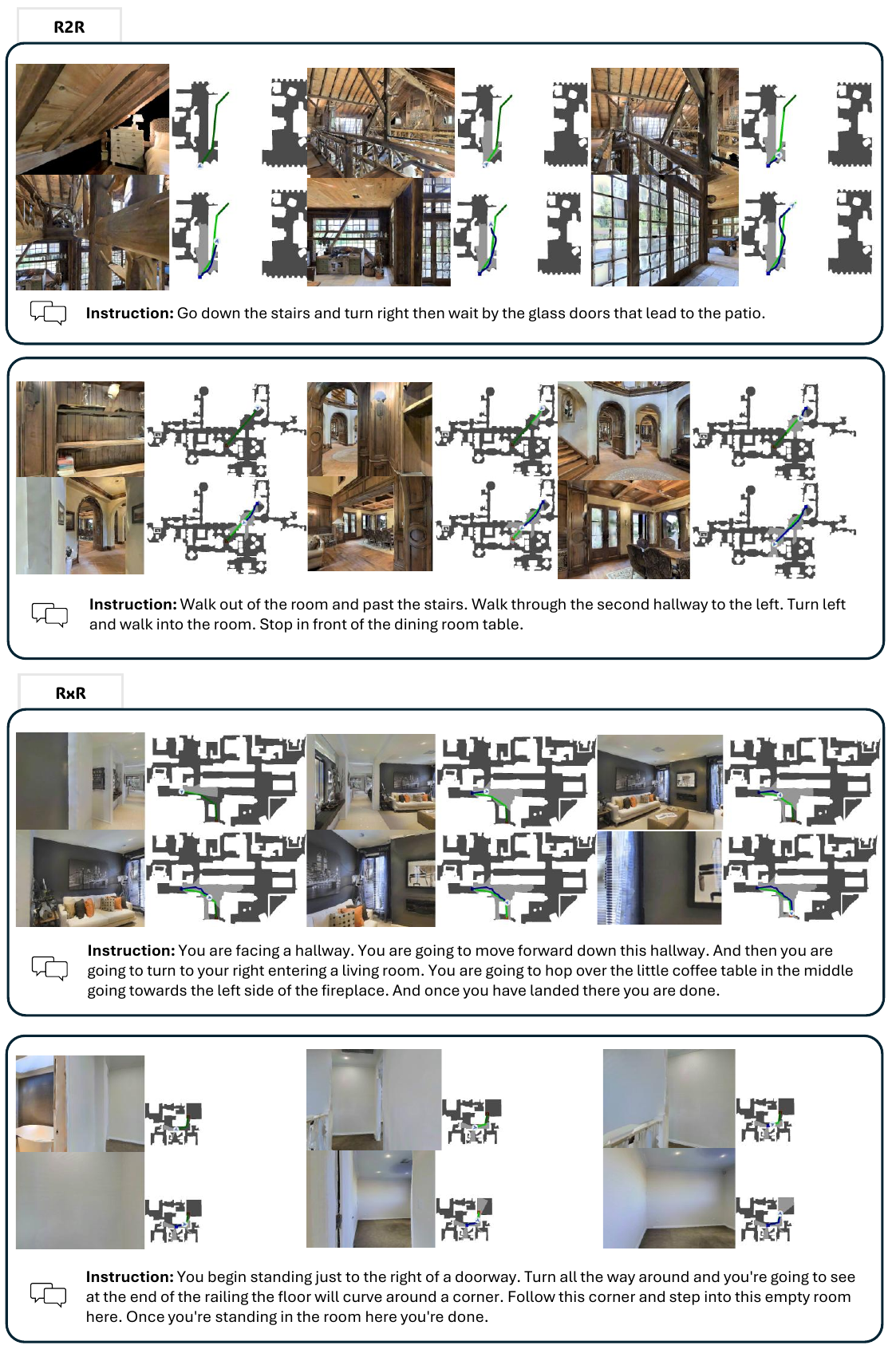}
    \caption{Visualization of navigation trajectories on
    R2R-CE and RxR-CE.}
    \label{fig:qualitative-navigation}
\end{figure}
\endgroup

\section{Limitations and Future Work}
\label{app:limit}

RSI-Nav's self-improvement is constrained by the capabilities of its current policy. Since the same model generates instructions and executes them for verification, the resulting supervision may reinforce existing biases. Instructions that are valid but difficult to execute may also be rejected, potentially favoring easier trajectories and limiting further improvement. Future work could explore difficulty-aware data selection, adaptive filtering, and complementary verification signals to retain challenging examples while controlling supervision quality.

Execution-based filtering also provides an incomplete measure of instruction quality. Reaching the intended destination efficiently does not necessarily imply that every landmark description or spatial relation is correct. Such semantic errors may persist in accepted examples and propagate through subsequent updates. Future work could combine trajectory-level verification with grounded semantic checks and targeted human evaluation to better assess instruction fidelity and understand how errors accumulate across self-improvement rounds.

\end{document}